\documentclass[fleqn,10pt]{wlscirep}
\usepackage[utf8]{inputenc}
\usepackage[T1]{fontenc}
\usepackage{lineno}
\usepackage{tikz}
\usepackage{booktabs}
\usepackage{graphicx}
\usepackage{subcaption}

\usepackage{hyperref}
\usepackage{tablefootnote} 
\usepackage{multirow}

\title{Visual WetlandBirds Dataset: Bird Species Identification and Behavior Recognition in Videos}
\author[1]{Javier Rodriguez-Juan}
\author[1]{David Ortiz-Perez}
\author[1]{Manuel Benavent-Lledo}
\author[1]{David Mulero-Perez}
\author[1]{Pablo Ruiz-Ponce}
\author[2]{Adrian Orihuela-Torres}
\author[1,*]{Jose Garcia-Rodriguez}
\author[2,3]{Esther Sebastián-González}

\affil[1]{Department of Computer Technology, University of Alicante, Alicante, 03690, Spain}
\affil[2]{Department of Ecology, University of Alicante, Alicante, 03690, Spain}
\affil[3]{‘Ramón Margalef’ Multidisciplinary Institute for the study of the Environment. University of Alicante, Alicante, 02690, Spain}

\affil[*]{corresponding author(s): Jose Garcia-Rodriguez (jgarcia@dtic.ua.es)}

\begin{abstract}
The current biodiversity loss crisis makes animal monitoring a relevant field of study. In light of this, data collected through monitoring can provide essential insights, and information for decision-making aimed at preserving global biodiversity. Despite the importance of such data, there is a notable scarcity of datasets featuring videos of birds, and none of the existing datasets offer detailed annotations of bird behaviors in video format. In response to this gap, our study introduces the first fine-grained video dataset specifically designed for bird behavior detection and species classification. This dataset addresses the need for comprehensive bird video datasets and provides detailed data on bird actions, facilitating the development of deep learning models to recognize these, similar to the advancements made in human action recognition. The proposed dataset comprises 178 videos recorded in Spanish wetlands, capturing 13 different bird species performing 7 distinct behavior classes. In addition, we also present baseline results using state of the art models on two tasks: bird behavior recognition and species classification.
\end{abstract}
\begin{document}

\flushbottom
\maketitle

\thispagestyle{empty}


\section*{Background \& Summary}

Under the current scenario of global biodiversity loss, there is an urgent need for more precise and informed environmental management \cite{back:riordan}. In this sense, data derived from animal monitoring plays a crucial role in informing environmental managers for species conservation \cite{back:nichols-monitoring,back:hais}. Animal surveys provide important data on population sizes, distribution and trends over time, which are essential to assess the state of ecosystems and identify species at risk \cite{back:margules,back:smallwood}. By using systematic monitoring data on animal and bird populations, scientists can detect early warning signs of environmental changes, such as habitat loss, climate change impacts, and pollution effects \cite{back:morrison,back:bonebrake,back:carvalho}. This information helps environmental managers develop targeted conservation strategies, prioritize resource allocation, and implement timely interventions to protect vulnerable species and their habitats \cite{back:nichols-monitoring,back:joseph,back:nuttall}. Furthermore, bird surveys often serve as indicators of local ecological conditions, given birds' sensitivity to environmental changes, making them invaluable in the broader context of biodiversity conservation and ecosystem management \cite{back:fraixedas}. However, monitoring birds, as any other animal, is highly resource-consuming. Thus, automated monitoring systems that are able to reduce the investment required for accurate population data are much needed.

The first step to create algorithms that detect species automatically is to create datasets with information on the species traits to train those algorithms. For example, a common way to classify species is by their vocalizations \cite{back:bird-vocaliz}. For this reason, organizations such as the Xeno-Canto Foundation \footnote{\url{https://xeno-canto.org/}} compiled a large-scale online database \cite{back:xeno-canto} of bird sounds from more than 200,000 voice recordings and 10,000 species worldwide. This dataset was crowsourced and today it is still growing. The huge amount of data provided by this dataset has facilitated the organization of challenges to create bird-detection algorithms using acoustic data in understudied areas, such as those led by Cornell Lab\footnote{\url{https://www.birds.cornell.edu/home/}}. This is the case of BirdCLEF2023\cite{back:african-clef}, or BirdCLEF2024 \cite{back:indian-clef}, which used acoustic recordings of eastern African and Indian birds, respectively. While these datasets contain many short recordings from a wide variety of different birds, other authors have released datasets composed of fewer but longer recordings, which imitate a real wildlife scenario. Examples of this are NIPS4BPlus \cite{back:nips}, which contains 687 recordings summing a total of 30 hours of recordings or BirdVox-full-night \cite{back:birdvox}, which has 6 recordings of 10 hours each.

Although audio is a common way to classify bird species and the field of bioacoustics has increased tremendously in the latest years, another possible approach to identify species automatically is using images \cite{back:bird-images}. One of such bird image datasets is Birds525,\footnote{\label{footnote:birds525}\url{https://www.kaggle.com/datasets/gpiosenka/100-bird-species}}, which offers a collection of almost 90,000 images involving 525 different bird species. Another standard image dataset is CUB-200-2011 \cite{back:cub}, which provides 11,788 images from 200 different bird species. This dataset not only provides bird species, but also bounding boxes and part locations for each image. There are also datasets aimed at specific world regions like NABirds \cite{back:nabirds}, which includes almost 50,000 images from the 400 most common birds seen in North America. This dataset provides a fine-grained classification of species as its annotations differentiate between male, female and juvenile birds. These datasets can be used to create algorithms for the automatic detection of the species based on image data.

However, another important source of animal ecology information that has been much less studied because of the technological challenges of its use are videos. Video recordings may offer information not only about which species are present in a specific place, but also about their behavior. Information about animal behavior may be very relevant to inform about individual and population responses to anthropogenic impacts and has therefore been linked to conservation biology and restoration success \cite{alados1996fractal,lindell2008value,berger2016systematic,goldenberg2017challenges}. Besides its potential for animal monitoring and conservation, the number of databases on wildlife behavior are more limited. For example, the VB100 dataset \cite{back:vb100}, comprises 1416 clips of approximately 30 seconds. This dataset involves 100 different species from North American birds. The unique dataset comprised by annotated videos with birds behavior available in the literature is the Animal Kingdom dataset \cite{back:ak}, which is not specifically aimed at birds and contains annotated videos from multiple animals. Specifically, it contains 30,000 video sequences of multi-label behaviors involving 6 different animal classes. Table \ref{tab:data-summary} summarizes the main information of the datasets reviewed.

\def\checkmark{\tikz\fill[scale=0.4](0,.35) -- (.25,0) -- (1,.7) -- (.25,.15) -- cycle;}
\begin{table}[htbp]
    \centering
    \begin{tabular}{c|c|cccc}
        \hline
        Name & Modality & Region & Samples & Species & Only birds\\
        \hline{1pt}
        Xeno-canto~\cite{back:xeno-canto} & \multirow{5}{*}{Audio} & All & +200,000 \tablefootnote{\label{footnote:table}As it is a crowdsourced project, it grows with the time} & 12115$^{ \ref{footnote:table}}$ & \checkmark\\
        BirdCLEF2023~\cite{back:african-clef} & & Eastern Africa & 16,900 & 264 & \checkmark\\
        BirdCLEF2024~\cite{back:indian-clef} & & India & 24,460 & 942 & \checkmark\\
        NIPS4BPlus~\cite{back:nips} & & Spain/France & 687 & 61 & \checkmark\\
        BirdVox-full-night~\cite{back:birdvox} & & USA & 6 & 25 & \checkmark\\
        \cline{1-6}
        Birds525$^ {\ref{footnote:birds525}}$ & \multirow{3}{*}{Image} & All & 89,885 & 525 & \checkmark\\
        CUB-200-2011~\cite{back:cub} & & All & 11,788 & 200 & \checkmark\\
        NABirds~\cite{back:nabirds} & & North America & 48,000 & 400 & \checkmark\\
        \cline{1-6}
        VB100~\cite{back:vb100} & \multirow{3}{*}{Video} & North America & 1,416 & 100 & \checkmark\\
        Animal Kingdom~\cite{back:ak} & & All & 30,000 & - & \\
        WetlandBirds (\emph{Proposed}) & & Spain & 178 & 13 & \checkmark \\
        \hline
    \end{tabular}
    \caption{Summary of reviewed bird datasets.}
    \label{tab:data-summary}
\end{table}

Due to the scarcity of datasets involving birds videos annotated with its behaviors, this study proposes the development of the first fine-grained behavior detection dataset for birds. Differently from Animal Kingdom, where a video is associated with the multiple behaviors happening, in our dataset, spatio-temporal behavior annotations are provided. This implies that videos are annotated per-frame, where the behavior happening and the location is annotated in each frame (\emph{i.e.} bounding box). Moreover, the identification of the bird species appearing in the video is also provided. The proposed dataset is composed by 178 videos recorded in Spanish wetlands, more specifically in the region of Alicante (southeastern Spain). The 178 videos expand to 2765 behavior clips involving 13 different bird species. The average duration of each of the behavior clips is 19.84 seconds and the total duration of the dataset recorded is 58 minutes and 53 seconds. The annotation process involved several steps of data curation, with a technical team working alongside a group of professional ecologists.

Table \ref{fig:species-annotation} reflects the different species collected for the dataset, distinguishing between their common and scientific names. The number of videos and minutes recorded for each species is also included.


\begin{table}[htbp]
\centering
\begin{tabular}{cccc}
\hline
Common name          & Scientific name                     & Videos & Recorded minutes \\ \hline
Yellow-legged Gull   & \textit{Larus michahellis}          & 13     & 5.08             \\
White wagtail        & \textit{Motacilla alba}             & 13     & 4.33             \\
Squacco Heron        & \textit{Ardeola ralloides}          & 15     & 4.94             \\
Northern shoveler    & \textit{Spatula clypeata}           & 14     & 3.49             \\
Mallard              & \textit{Anas platyrhynchos}         & 10     & 2.94             \\
Little-ringed plover & \textit{Charadrius dubius}          & 10     & 1.93             \\
Glossy ibis          & \textit{Plegadis falcinellus}       & 8      & 3.96             \\
Gadwall              & \textit{Mareca strepera}            & 13     & 2.59             \\
Eurasian moorhen     & \textit{Gallinula chloropus}        & 18     & 9.18             \\
Eurasian magpie      & \textit{Pica pica}                  & 16     & 5.95             \\
Eurasian coot        & \textit{Fulica atra}                & 19     & 4.11             \\
Black-winged stilt   & \textit{Himantopus himantopus}      & 14     & 3.55             \\
Black-headed gull    & \textit{Chroicocephalus ridibundus} & 15     & 6.84             \\ \hline
\end{tabular}
\caption{Data for each of the annotated species.}
\label{fig:species-annotation}
\end{table}

\begin{figure}[htbp]
  \centering
  \includegraphics[width=0.8\linewidth]{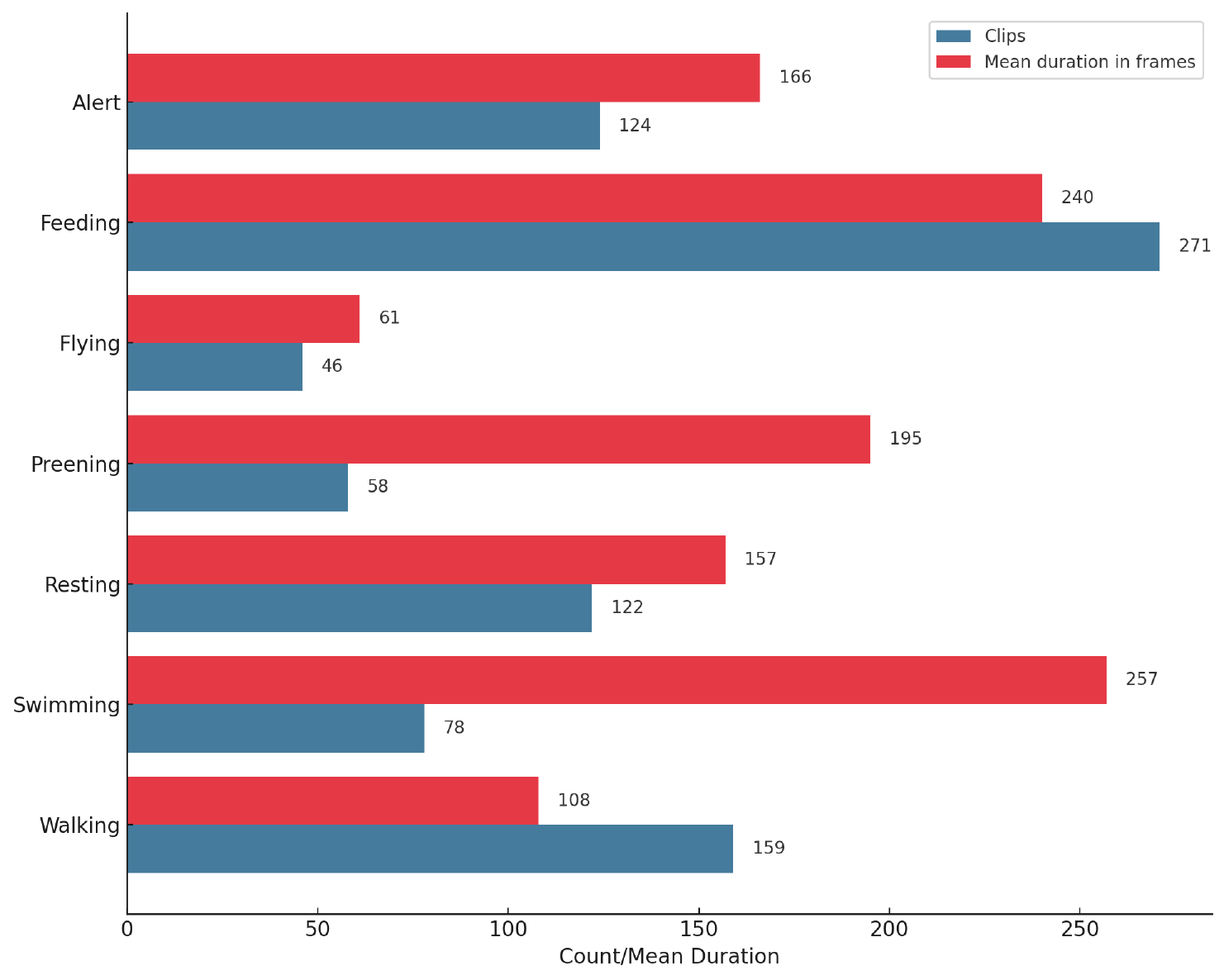}
  \caption{Comparison of total number of clips per behavior and mean duration of behavior clips.}
  \label{fig:clips-stats}
\end{figure}

Seven main behaviors were identified as key activities recorded in our dataset. These represent the main activities performed by waterbirds in nature \cite{back:rose}. In Figure \ref{fig:clips-stats}, these behaviors are specified alongside the number of clips recorded per each of them and the mean duration of each behavior in frames. A clip is a piece of video where a bird is performing a specific behavior. Animals often change among actions very fast, as a response to the changing environment. Thus, to consider a collection of movements of a bird as a behavior, this had to last a minimum of 30 frames, otherwise this collection of movements was identified as a sub-movement of another main behavior, which is the one annotated for those frames. 

\begin{figure}[htbp]
    \centering
    \includegraphics[width=0.8\linewidth]{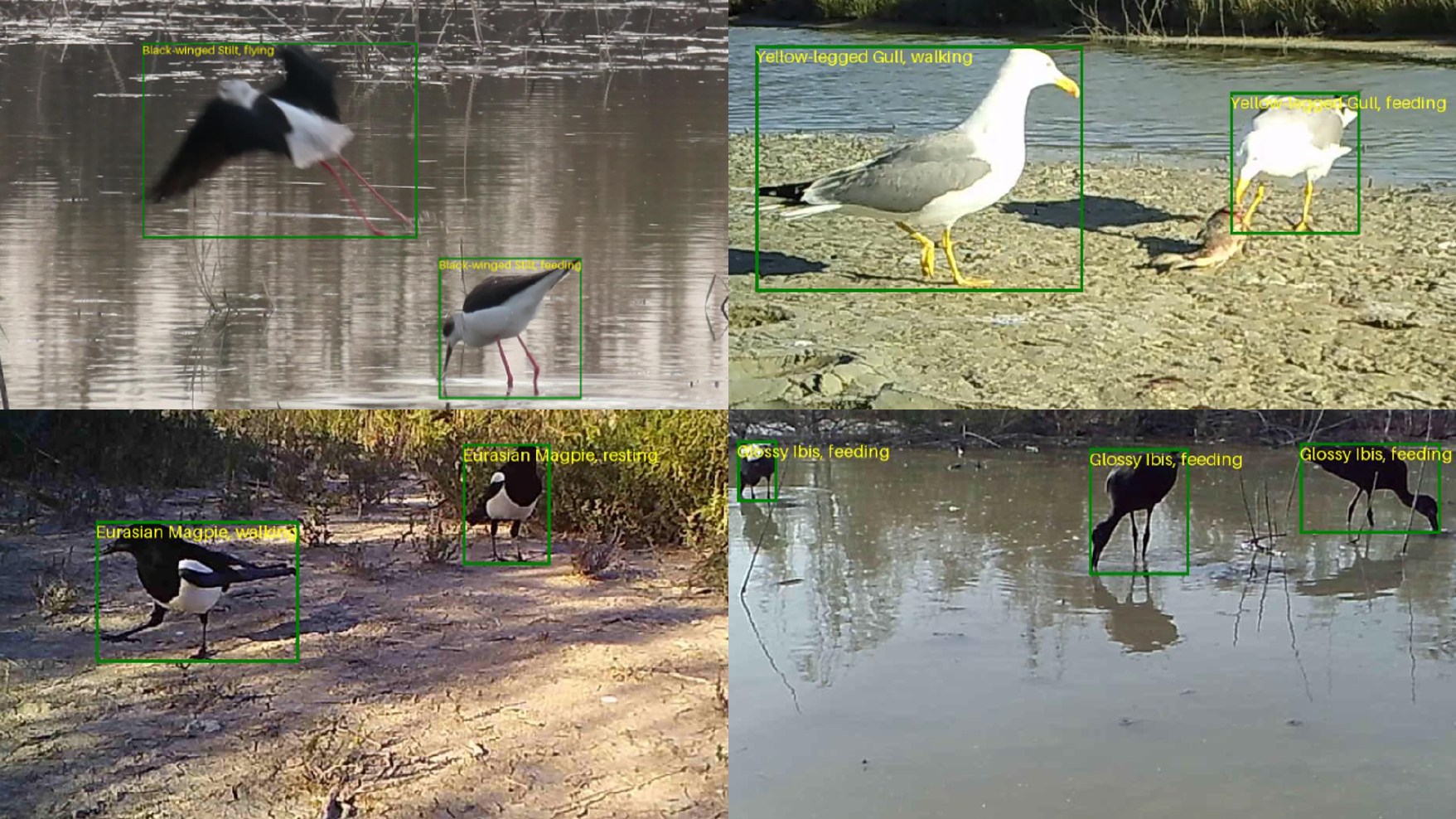}
    \caption{Frame samples where gregarious birds appear performing different behaviors.}
    \label{fig:gregarious-birds}
  \end{figure}

This dataset contains not only videos with a single individual, but also videos where several bird individuals appear together. This is the case for gregarious species, which are species that concentrate in an area for the purpose of different activities. Although the individuals of gregarious species often share the same behavior at the same time, it is also common that several behaviors can be seen in the same video at the same time. Figure \ref{fig:gregarious-birds} shows some samples of videos where this happens. Videos involving different birds and/or performing different activities sequentially were cut in clips where a unique individual is performing a unique behavior in order to get the statistics shown in Figure \ref{fig:clips-stats}. 

Between the seven behaviors proposed, it should be underlined the difference between the \emph{Alert}, \emph{Preening} and \emph{Resting} behaviors. These action distinctions were established by the ecology team. We considered that the bird was \emph{Resting} when it was standing without making any movement. The bird was performing the \emph{Alert} behavior when it was moving its head from one side to another, moving and looking around for possible dangers. Finally, we considered that the bird was \emph{Preening} when it was standing and cleaning its body feathers with its beak. The remaining behaviors are not explained because of their obvious meaning.

As it can be seen in Figure \ref{fig:clips-stats}, the number of clips per behavior is unbalanced between classes. This is because the recording of videos where some specific behaviors are happening is more uncommon, as happens with \emph{Flying} or \emph{Preening}, which represent the activities with the lowest number of clips in the dataset. These behaviors are difficult to record since they are performed with a lower frequency. In order to be able to collect more data on these less common behaviors, more hardware and human resources (\emph{i.e.} cameras and professional ecologists) are needed to cover a wider area of the wetlands. Although the unbalanced nature of the behaviors, no balancing technique over this data was applied in the released dataset in order to maximize the number of different environments captured, ensuring in this way the variability of contexts where the birds are recorded.

Additionally, Figure \ref{fig:clips-stats} also shows the mean duration of the clips per behavior. It is worth noting the difference in the number of frames between \emph{Flying}, that represents the minimum with 61 frames with respect to \emph{Swimming}, which represents the absolute maximum with a value of 257 frames. This difference is explained in the nature of the behaviors, as swimming is naturally a slow behavior, which can be performed for a long time over the same area. However, flying is a fast behavior, and the bird quickly get outs of the camera focus, especially for videos obtained by camera traps, which cannot follow the bird while it is moving.



It is also common that birds perform two activities simultaneously. In that case, the most relevant behavior for the bird was the one annotated. In the proposed dataset, \emph{Feeding} behavior is the one which is commonly done simultaneously to others such as \emph{Walking} or \emph{Swimming}. As \emph{Feeding} behavior was considered by the ecologist experts as more relevant for birds, this behavior was always annotated in these cases. \\

\begin{figure}[htbp]
    \centering
    \includegraphics[width=0.8\linewidth]{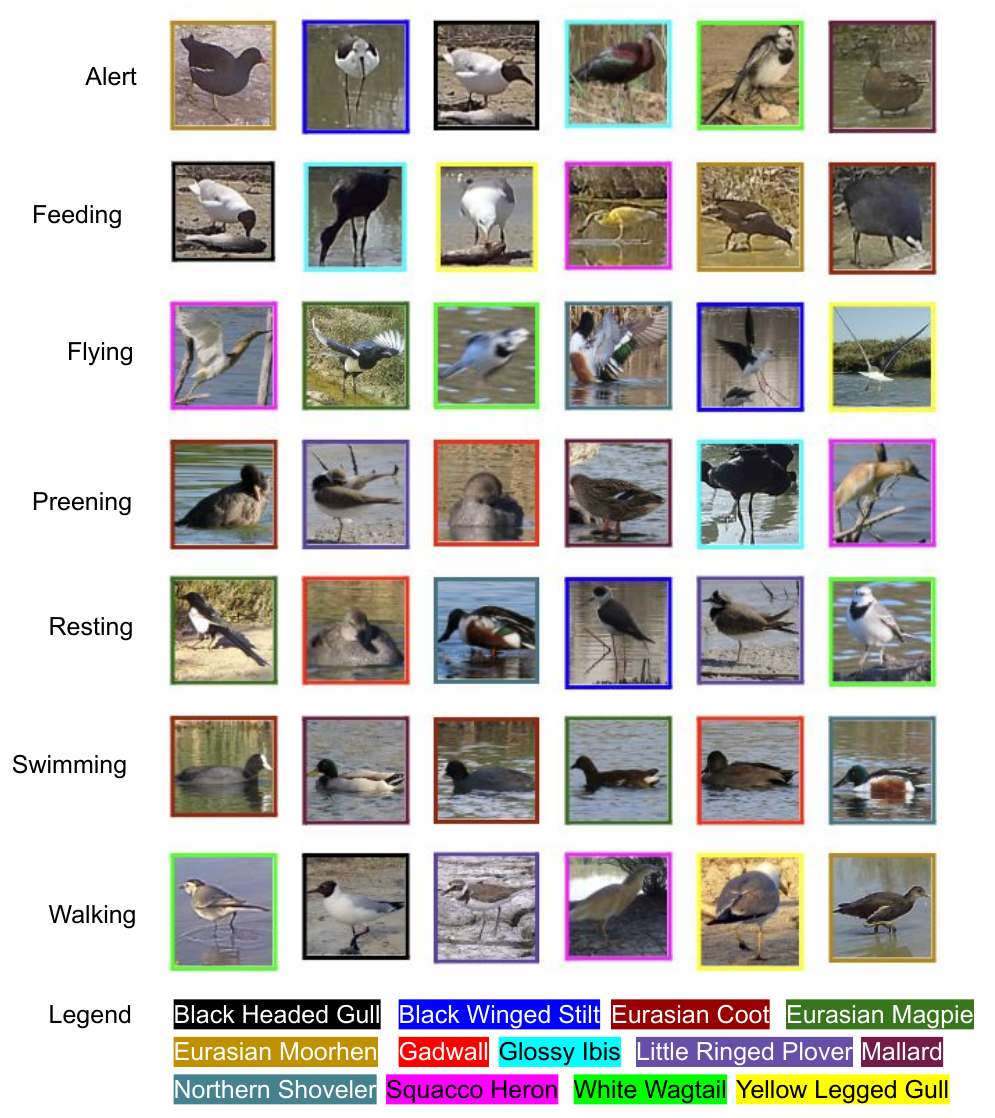}
    \caption{Video frame crops of bird species performing the 7 behaviors composing the dataset.}
    \label{fig:dataset-samples}
  \end{figure}

In order to collect the videos, we deployed a set of camera traps and high quality cameras in Alicante wetlands. The camera traps were able to automatically record videos based on the motion detected in the environment. We complemented the camera trap videos with recordings from high quality cameras. In these videos, a human is controlling the focus of the camera, obtaining better views and perspectives of the birds being recorded. Species recorded, behaviors identified and the camera deployment areas were described by professional ecologist based on their expertise. In Figure \ref{fig:dataset-samples} some video frame crops can be observed, where all the bird species developing the different behaviors available in the dataset can be seen.

After the data collection, a semi-automatic annotation method composed by an annotation tool and a deep learning model was used in order to get the videos annotated. After the annotation, a cross-validation was conducted to ensure the annotation quality. This method is deeply explained in the next section.

In order to test the dataset for species and behavior identification, two baseline experimentation were carried out: one for the for the behavior detection task, which involves the correct classification of the behavior being performed by one bird during a set of frames, and a second one for the bird classification task, which involves the classification of the specie and the correct localization of the bird given input frames.

\section*{Methods}

\subsection*{Data acquisition}

The acquisition of the data was conducted within Alicante wetlands, specifically within the wetlands of \emph{La Mata Natural Park} and \emph{El Hondo Natural Park} (sutheastern Spain). In these places, we deployed a collection of high-resolution cameras and camera traps in different areas of the wetlands. These areas were determined by the species expected to be recorded, as different species can be commonly seen in different wetland areas. 

Camera traps are activated when movement is detected and thus can record for long periods of time without human intervention. The usage of automatic camera traps \cite{method:camera-trap1,method:camera-trap2,method:camera-trap3} is common in the monitoring of wildlife as it provides a low-cost approach to collect video and image data from the environment. However, the focus of this camera is fix and thus the videos of the same individual are often short. Manual cameras require the presence of a human while recording and are thus more time-consuming. Also, the presence of the cameraman may affect the animal behavior. However, it permits manual changes of cameras' perspectives in order to correctly record the bird behavior. As different cameras were used videos of different resolutions were obtained: 87 videos at 1920x1080px, 75 videos at 1296x720px, 14 videos at 1280x720px, 1 video at 960x540px and 1 video at 3840x2160px.


The species selected were the most common found in the wetlands of Alicante, facilitating the recording of videos and providing valuable data to the natural parks where videos were recorded. In terms of behaviors, we identified the most representative ones of the selected species, in order to cover as much as possible the range of activities developed by the birds.

\subsection*{Data annotation}

Accurate annotation of the captured data is a determining factor in obtaining relevant results when training deep learning models on this data. To ensure annotation accuracy, the use of annotation tools \cite{method:annot-tool1,method:annot-tool2} has been extended, as they provide a user-friendly interface that makes this process easy and accessible to non-technical staff.

There are many open-source annotation tools available on the market. CVAT\footnote{\url{https://github.com/cvat-ai/cvat}} is one of the most popular ones, as it provides annotation support for images and videos, including a variety of formats for exporting the data. VoTT \footnote{\url{https://github.com/microsoft/VoTT}} is also popular when annotating videos, as it offers multiple annotation shapes and integration with Microsoft services to easily upload data to Azure.\footnote{\url{https://azure.microsoft.com/}} Other simpler annotation tools are labelme\footnote{\url{https://github.com/labelmeai/labelme}} or LabelImg,\footnote{\url{https://github.com/HumanSignal/labelImg}} which are aimed at annotating images and their capabilities are more limited. For our purpose, we decided to use CVAT because of the large number of exportable formats available, the great collaborative environment it offers, and its easy integration with semi-automatic and automatic annotation processes.

As the need for larger amounts of data to train deep learning models increased, researchers began to enhance annotation tools with automatic systems that could alleviate this task. Annotation tools integrate machine learning models \cite{method:automated-cvat} that can automatically infer what would otherwise be manually annotated. Common tasks performed by automated annotation tools are object detection \cite{method:automated-od} and semantic segmentation.\cite{method:automated-ss} While the former predicts the bounding box and class of each object in the image, the latter predicts regions of interest associated with specific categories.

Although automated annotation systems have demonstrated strong performance, semi-automated annotation processes are ultimately used because they ensure the creation of highly accurate annotations while greatly reducing the amount of human intervention required. Semi-automated annotation studies are widely used in the medical field,\cite{method:automated-medical,methods:automated-medical-2} where precision is a key factor throughout the design.

\begin{figure}[htbp]
    \centering
    \includegraphics[width=1\linewidth]{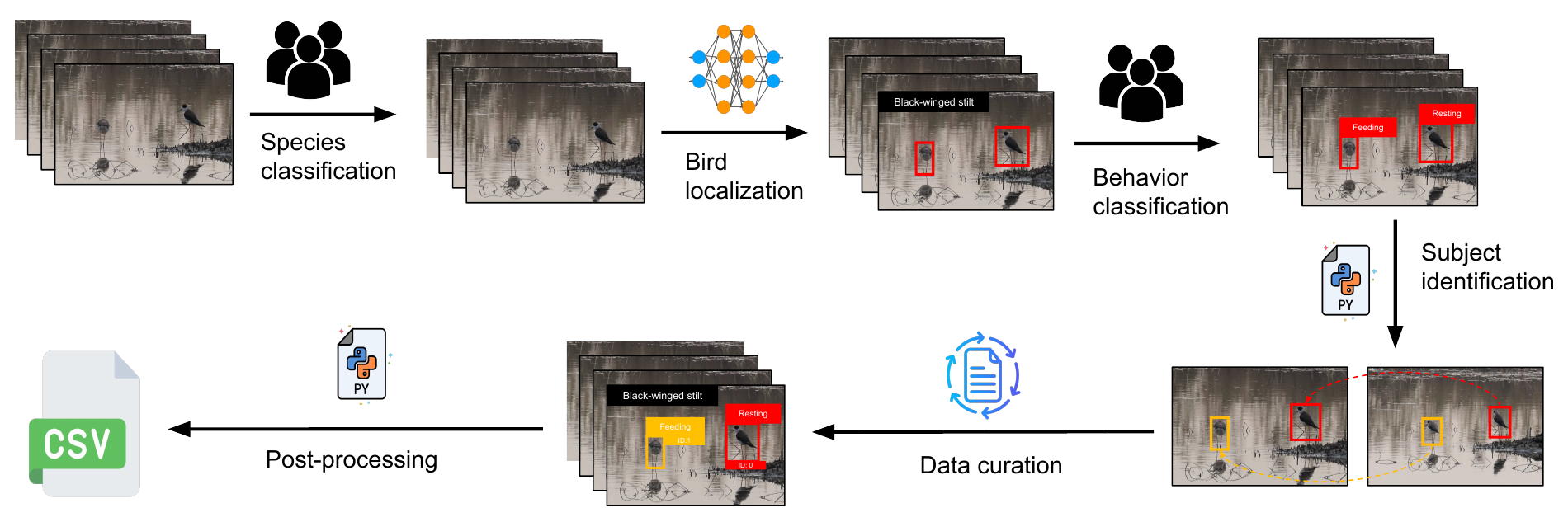}
    \caption{Visual representation of stages involved in the annotation process. Birds are first classified into species by annotators and localized using a YOLO model, then annotators recognize bird behaviors and subjects are identified using a Python script, and finally the data is curated and post-processed.}
    \label{fig:annot-process}
  \end{figure}

In this study, a semi-automated annotation approach was followed, based on CVAT and its possible integration with powerful computer vision models. Our approach consisted of five main steps: Species classification, bird localization, behavior classification, subject identification, data curation, and post-processing. Each of these stages is described in more detail below. Figure \ref{fig:annot-process} shows this process.

\begin{enumerate}
    \item \textbf{Species classification:} In this first step, the ecologists labeled each video with the main bird species that appeared. The main species is that of the bird in the focus of the camera. This way, annotations of birds that are different from the main species will not be included in the video annotations.

    \item \textbf{Bird localization:} Then, an object detection model is used to predict the localization of the bounding boxes of the birds that appear in each of the video frames. For ease of implementation, YOLOv7 \cite{method:yolo} was chosen as the object detection model because it is predefined integration into CVAT. Since the model provided by CVAT is trained on general purpose data, the class predicted by default for each bounding box is not be the bird species, but the class \emph{bird}. To avoid manually changing all the bounding box classes, we used an option provided by CVAT to associate a user-defined class with the class detected by the model. In this way, the class \emph{bird} was associated with the species appearing in the video. Since each video is restricted to having only one species of bird, it is possible to associate \emph{bird} with a specific species.
    
    \item \textbf{Behavior classification:} After automatically annotating the bounding boxes, our team of ecologists performed the second step of the annotation process, which is twofold. First, they checked and corrected erroneous bounding boxes, and second, they annotated for each bounding box the behavior performed by each bird. To annotate the behaviors, CVAT bounding boxes \emph{tags} were used.
    
    \item \textbf{Subject identification:} When using automatic annotation models such as YOLOv7, CVAT does not support bounding box correspondence between frames. In other words, if a video shows two birds developing different behaviors, there is no relationship between the bounding boxes of adjacent frames, so it is not possible to analyze the birds' behaviors. This is not possible because the next frame will show two new bounding boxes whose relation to the one being analyzed is not known. To solve this problem, the Euclidean distance \cite{records:euclidean} was used to correlate bounding boxes of adjacent frames. The euclidean distance calculates the distance between the centers of the bounding boxes of adjacent frames and then correlates the bounding boxes with the minimum distance. The center of the bounding box was calculated as follows:

    \begin{equation}
    c = \left( \frac{x_{\min} + x_{\max}}{2}, \frac{y_{\min} + y_{\max}}{2} \right)
    \end{equation}
    
    Given the centre of the bounding boxes, the Euclidean distance was calculated as:
    
    \begin{equation}
    d(c_1, c_2) = \sqrt{(x_2 - x_1)^2 + (y_2 - y_1)^2}
    \end{equation}
    \begin{align*}
    \text{where } c_1 &= (x_1, y_1) \\
    \text{and } c_2 &= (x_2, y_2)
    \end{align*}

    \item \textbf{Data curation:} After the labeling of species, bounding boxes, behaviors, and subjects, an overall review of all annotations was conducted to ensure the high quality of the data. To conduct the review, videos were assigned to all ecologists equally.
    
    \item \textbf{Post-processing:} Once the annotations were complete, their format was adapted to make them easy to use and understand. To achieve this goal, the approach used in the AVA-Kinetics dataset \cite{records:ava} was followed. In this approach, a CSV file was used to contain annotations containing localized behaviors of multiple subjects. To export the data into the CSV format, the data was first exported from CVAT using the CVAT Video 1.1 format. Some Python scrips were then used to extract only the relevant information from the exported data and dump it into the output CSV file.
    
\end{enumerate}

\section*{Data Records}

The dataset presented in this study is open access and accessible through Zenodo\footnote{\url{https://zenodo.org/records/14355257}}. Within this Zenodo repository, there are four main elements:

\begin{itemize}
    \item \textbf{Videos folder:} This folder contains the 178 videos that comprise the dataset. Videos are identified by their name, which is composed of a numeric value and the species that appears in the video. The format is the following "ID-VIDEO.SPECIES-NAME.mp4".
    
    
    \item \textbf{Bounding boxes CSV:} The \emph{bounding\_boxes.csv} file contains all the annotations of the data set. It follows a format of 10 columns, ordered as follows: Global identifier of the row, video identifier, frame identifier within the video, activity identifier, subject identifier, species appearing in the video, and the four coordinates of the bounding box (top-left x-coordinate, top-left y-coordinate, bottom-right x-coordinate, and bottom-right y-coordinate). Each of the CSV rows represents the information of one bounding box within one frame of a video.
    
    \item \textbf{Behavior identifiers CSV:} The \emph{behavior\_ID.csv} file contains a mapping of the seven behavior classes that make up the data set and their numeric identifiers.
    
    \item \textbf{Species identifiers CSV:} The file \emph{species\_ID.csv} contains a mapping between the 13 different bird species and their numerical identifier.

    \item \textbf{Splits JSON:} The \emph{splits.json} file contains the videos associated with each train, validation and test split.
\end{itemize}



\section*{Technical Validation}


To ensure high quality recordings and accurate annotations, the entire process was carried out by expert ecologists, as mentioned in the Data Annotation section. Firstly, video recordings were supervised by a group of experts who set up camera traps in strategic areas and also manually recorded some high-quality videos. For each video, these experts annotated the species appearing in the video. The same experts then corrected bounding box errors and annotated bird behavior, together with a number of collaborators with a background in ecology. Finally, a final stage of cross-checking of annotations was carried out by the experts and collaborators. The expertise of the annotators responsible for collecting and annotating the videos, together with the final cross-review process, ensures the quality and cleanliness of the data.

As this dataset has been conceived mainly to be used in deep learning pipelines, baseline pipelines will be given to demonstrate the applicability of the data presented in this work within deep learning workflows. As mentioned previously, the purpose of this dataset is twofold, as it provides annotation data for performing bird species detection and behavior classification tasks. Thus, one baseline per each task was developed. PyTorch\footnote{\url{https://pytorch.org}} was used in both cases as coding platform.

\subsection*{Species classification}

First, a baseline pipeline for species classification was developed. This baseline is based on a YOLOv9 \cite{tech:yolov9} model trained over 50 epochs in the proposed dataset.

For efficient training, a downsampling of 10 is performed on the frames extracted from the videos. This can be done without affecting the performance of the model, as the difference between successive frames is minimal. The frames were extracted while maintaining the source FPS (Frames Per Second) of each video. During the training stage, a learning rate of 0.01 was used and a GeForce RTX 3090 GPU was used as the hardware platform. The test results from the baseline are shown next:

\begin{table}[htpb]
\centering
\begin{tabular}{lcccc}
\hline
Model  & Precision & Recall & mAP50 & mAP50-95 \\ \hline
YOLOv9 & 0.835     & 0.759  & 0.801 & 0.556 \\ \hline
\end{tabular}
\caption{Results of the YOLO-based baseline developed for bird species classification.}
\label{tab:spc-clf-results}
\end{table}

The table \ref{tab:spc-clf-results} shows the test results for species classification in terms of precision, recall, mAP50 and mAP50-95 metrics. mAP50-95 is a common object recognition metric that refers to the mAP (mean Average Precision) computed over 10 different IoU (Intersection over Union) thresholds, specifically from 0.50 to 0.95 in increments of 0.05. The results show how YOLOv9 achieves a strong performance for the task, with a maximum precision of 0.835, while maintaining a high recall of 0.759. The mAP metrics, which measure the accuracy of bounding box localizations, are also strong, reaching 0.801 for mAP50 and 0.556 for mAP50-95, which is a good result considering the challenge of obtaining strong mAP scores when high IoU values are used.

\begin{figure}[htbp]
  \centering
  \includegraphics[width=0.8\linewidth]{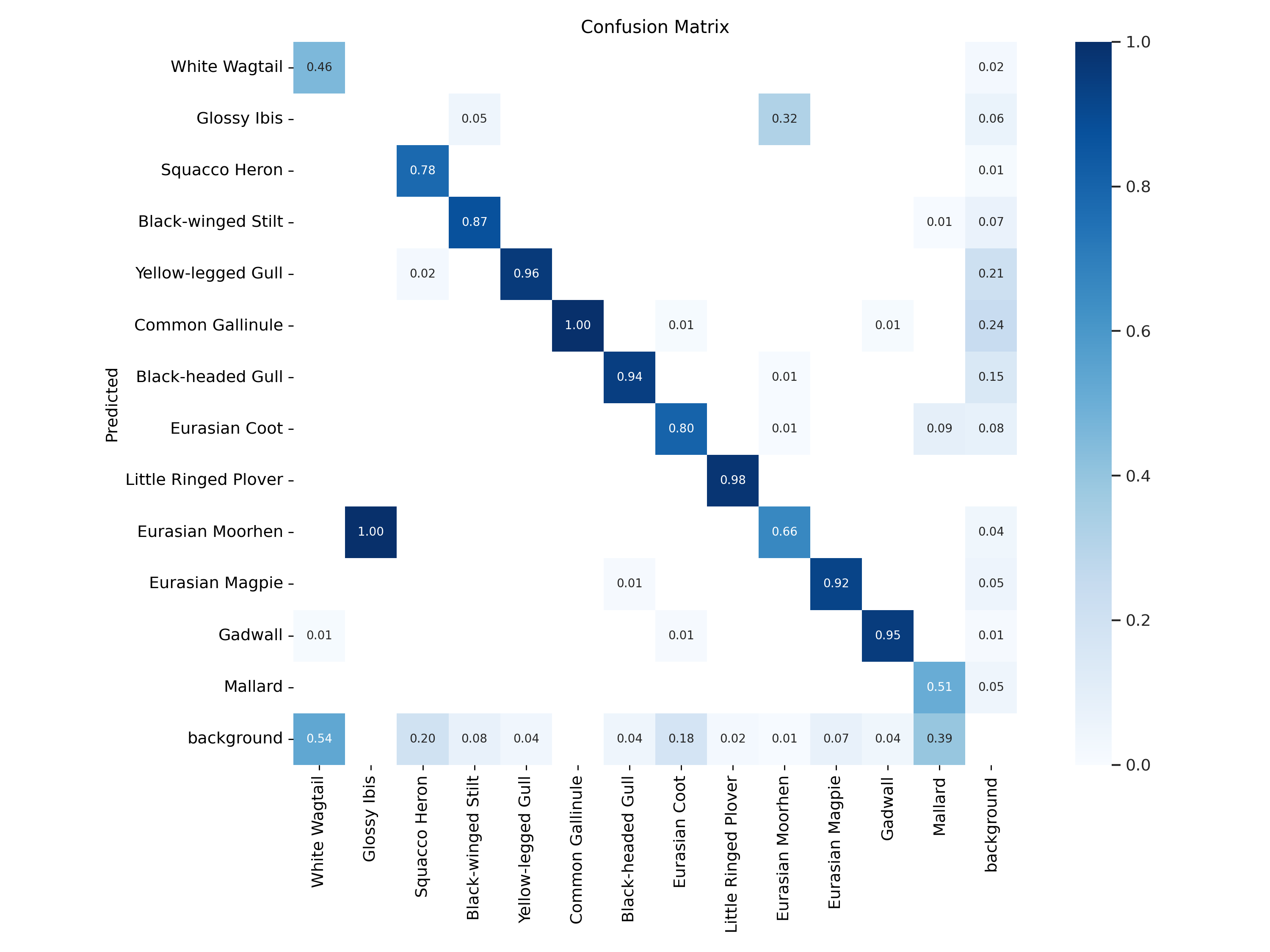}
  \caption{Confusion matrix of species classification pipeline.}
  \label{fig:cf-spc-clf}
\end{figure}

To provide a more comprehensive understanding of the evaluation, the confusion matrix for the results is given. Figure \ref{fig:cf-spc-clf} shows the confusion matrix, where it can be observed that the majority of the errors are due to the confusion of the ground truth class with the background class.

\subsection*{Behavior detection}

Secondly, the behavior detection baseline is presented. In this baseline, four different video classification models were trained end-to-end to perform the behavior classification task. The trained models were Video MViT,\cite{tech:mvit} Video S3D,\cite{tech:s3d} Video SwinTransformer \cite{tech:swin} and Video ResNet.\cite{tech:vresnet} All model architectures and pretrained weights were extracted from PyTorch. 

For the training stage, the input videos were downsampled with a downsample rate of 3, selecting the first frame as the representative of each set (\emph{i.e.} only the first frame of each set of 3 is kept). Train, test and validation splits were generated from the full set of videos with a distribution 70-15-15. The splits were constructed using a stratified strategy based on the species and behaviors appearing in the videos. The distribution was computed taking into account the number of frames which constitute each video (\emph{e.g.} 1 video with 1000 frames is equivalent to 5 videos with 200 frames). Regarding the training hyperparameters, a learning rate tuning was conducted using a uniform sampling strategy with minimum and maximum values of 0.0001 and 0.01, respectively. Similarly to the species classification baseline, training was performed on a GeForce RTX 3090 GPU. The results for each model are shown below:

\begin{table}[htpb]
\centering
\begin{tabular}{lcc}
\hline
Model           & Learning rate & Accuracy \\ \hline
MViT \cite{tech:mvit} & 0.005         & 0.51     \\
S3D \cite{tech:s3d}            & 0.005         & 0.29     \\
SwinTransformer \cite{tech:swin} & 0.009         & 0.51     \\
Video ResNet \cite{tech:vresnet}         & 0.003         & 0.56     \\ \hline
\end{tabular}
\caption{Results of the baseline models for behavior detection in terms of accuracy. The learning rate shown is the one that achieved the highest accuracies during hyperparameter tuning..}
\label{tab:behav-clf-results}
\end{table}

From the Table \ref{tab:behav-clf-results} it can be concluded  that the Video ResNet model is the one which learns better the complexity of the dataset, showcasing a maximum performance of 0.56. Conversely, the model with the lowest score is the S3D model, with an accuracy of 0.29. These results show the challenge posed by the dataset under study, which presents a limited amount of data. The limited amount of data available to train complex deep learning models demonstrates the need for more resources to capture more data. Furthermore, the development of new training strategies and deep learning architectures that fit the data needs should be explored in order to improve the baseline results obtained. 



\section*{Usage Notes}

Since the data annotations are provided in CSV format, it is recommended to use Python libraries such as Pandas,\footnote{\url{https://pandas.pydata.org/}} which is specifically designed to read and manage CSV data. In the GitHub repository containing the code, there are usage examples of how to load and prepare the data to be fed into deep learning models. It is recommended to read the \emph{dataset.py} script in the \emph{behavior\_detection} directory as an example.



\section*{Code availability}

The data processing and experimentation code shown in the Technical Validation section is available on GitHub\footnote{\url{https://github.com/3dperceptionlab/Visual-WetlandBirds}}. The GitHub repository is organized into two main directories. The \emph{species\_classification} directory contains all the code related to the species classification, and the \emph{behavior\_detection} directory contains the experiment with the behavior detection models proposed for the dataset.


\bibliography{sample}


\section*{Acknowledgements}
\sloppy
We would like to thank ”A way of making Europe” European Regional Development Fund (ERDF) and MCIN/AEI/10.13039/501100011033 for supporting this work under the “CHAN-TWIN” project (grant TED2021-130890B- C21 and HORIZON-MSCA-2021-SE-0 action number: 101086387, REMARKABLE, Rural Environmental Monitoring via ultra wide-ARea networKs And distriButed federated Learning. This work is part of the HELEADE project (TSI-100121-2024-24), funded by Spanish Ministry of Digital Processing and by the European Union NextGeneration EU. This work has also been supported by three Spanish national and two regional grants for PhD studies, FPU21/00414, FPU22/04200, FPU23/00532, CIACIF/2021/430 and CIACIF/2022/175.

\section*{Author contributions statement}

J.R.J. conceived the manuscript, J.R.J., D.O.P., M.B.L., D.M.P. and P.R.P. conducted the experiments and analysed the results, A.O.T. and E.S.G. managed the data collection and annotation, J.G.R. led the project management. All authors contributed to the annotation of the data and reviewed the manuscript.

\section*{Competing interests}

The authors declare no competing interests.



\end{document}